\documentclass[format=sigconf]{acmart}

\ccsdesc[500]{General and reference~Evaluation}
\ccsdesc[300]{Social and professional topics~User characteristics}
\ccsdesc[500]{Software and its engineering~Use cases}
\ccsdesc[300]{Software and its engineering~Documentation}
\ccsdesc[300]{Software and its engineering~Software evolution}
\ccsdesc[300]{Human-centered computing~Walkthrough evaluations}

\keywords{datasheets, model cards, documentation, disaggregated evaluation, fairness evaluation, ML model evaluation, ethical considerations}

\usepackage{booktabs} 
\usepackage{framed} 
\usepackage{enumitem}
\usepackage{subcaption}
\usepackage{url}
 \usepackage{array,multirow,graphicx}
 \usepackage{float}

\setcopyright{rightsretained}

\acmYear{2019}\copyrightyear{2019}
\setcopyright{acmcopyright}
\acmConference[FAT* '19]{FAT* '19: Conference on Fairness, Accountability, and Transparency}{January 29--31, 2019}{Atlanta, GA, USA}
\acmBooktitle{FAT* '19: Conference on Fairness, Accountability, and Transparency, January 29--31, 2019, Atlanta, GA, USA}
\acmPrice{}
\acmDOI{10.1145/3287560.3287596}
\acmISBN{978-1-4503-6125-5/19/01}
\begin{document}
\title{Model Cards for Model Reporting}

\author{Margaret Mitchell, Simone Wu, Andrew Zaldivar, Parker Barnes, Lucy Vasserman, Ben Hutchinson, Elena Spitzer, Inioluwa Deborah Raji, Timnit Gebru}
\email{{mmitchellai,simonewu,andrewzaldivar,parkerbarnes,lucyvasserman,benhutch,espitzer,tgebru}@google.com}
\email{deborah.raji@mail.utoronto.ca}

\begin{abstract}
Trained machine learning models are increasingly used to perform high-impact 
tasks in areas such as law enforcement, medicine, education, and employment. In order to clarify the intended use cases of machine learning models and minimize their usage in contexts for which they are not well suited, we recommend that released models be accompanied by documentation detailing their performance characteristics. In this paper, we propose a framework that we call model cards, to encourage such transparent model reporting. Model cards are short documents accompanying trained machine learning models that provide benchmarked evaluation in a variety of conditions, such as across different cultural, demographic, or phenotypic groups (e.g., race, geographic location, sex, Fitzpatrick skin type \cite{fitzpatrick}) and intersectional groups (e.g., age and race, or sex and Fitzpatrick skin type) that are relevant to the intended application domains. Model cards also disclose the context in which models are intended to be used, details of the performance evaluation procedures, and other relevant information. While we focus primarily on human-centered machine learning models in the application fields of computer vision and natural language processing, this framework can be used to document any trained machine learning model. To solidify the concept, we provide cards for two supervised models: One trained to detect smiling faces in images, and one trained to detect toxic comments in text. We propose model cards as a step towards the responsible democratization of machine learning and related artificial intelligence technology, increasing transparency into how well artificial intelligence technology works.  We hope this work encourages those releasing trained machine learning models to accompany model releases with similar detailed evaluation numbers and other relevant documentation.
\end{abstract}

\maketitle

\section{Introduction}
Currently, there are no standardized documentation procedures to communicate the performance characteristics of trained machine learning (ML) and artificial intelligence (AI) models. This lack of documentation is especially problematic when models are used in applications that have serious impacts on people's lives, such as in health care \cite{nervana,innereye,digital_reasoning}, employment \cite{ideal,avrio,entelo}, education \cite{gooru,turnitin} and law enforcement \cite{FBI2012,garvie2016,ProPublica2016,ftc_2016}.

Researchers have discovered systematic biases in commercial machine learning models used for face detection and tracking \cite{Desi09,Wang2010,buolamwini2016ted}, attribute detection \cite{BuolamwiniGebru2018}, criminal justice \cite{COMPAS2016}, toxic comment detection \cite{DixonEtAl2018}, and other applications. However, these systematic errors were only exposed after models were put into use, and negatively affected users reported their experiences. For example, after MIT Media Lab graduate student Joy Buolamwini found that commercial face recognition systems failed to detect her face \cite{buolamwini2016ted}, she collaborated with other researchers to demonstrate the disproportionate errors of computer vision systems on historically marginalized groups in the United States, such as darker-skinned women \cite{BuolamwiniGebru2018,WakandaScorecard}. 
In spite of the potential negative effects of such reported biases,  
documentation accompanying trained machine learning models (if supplied) provide very little information regarding model performance characteristics, intended use cases, potential pitfalls, or other information to help users evaluate the suitability of these systems to their context. 
This highlights the need to have detailed documentation accompanying trained machine learning models, including metrics that capture bias, fairness and inclusion considerations. 

As a step towards this goal, we propose that released machine learning models be accompanied by short (one to two page) records we call model cards.  Model cards (for model reporting) are complements to ``Datasheets for Datasets'' \cite{GebruEtAl2018} and similar recently proposed documentation paradigms \cite{HollandEtAl2018,statements} that report details of the datasets used to train and test machine learning models. Model cards are also similar to the {\sc tripod} statement proposal in medicine \cite{tripod}.  We provide two example model cards in Section \ref{sec:examples}: A smiling detection model trained on the CelebA dataset \cite{LiuEtAl2015} (Figure \ref{fig:smiling_card}), and a public toxicity detection model \cite{perspective_api} (Figure \ref{fig:toxicity_card}). Where Datasheets highlight characteristics of the data feeding into the model, we focus on trained model characteristics such as the type of model, intended use cases, information about attributes for which model performance may vary, and measures of model performance. 

We advocate for measures of model performance that contain quantitative evaluation results to be broken down by individual cultural, demographic, or phenotypic groups, domain-relevant conditions, and intersectional analysis combining two (or more) groups and conditions. In addition to model evaluation results, model cards should detail the motivation behind chosen performance metrics, group definitions, and other relevant factors. Each model card could be accompanied with Datasheets \cite{GebruEtAl2018}, Nutrition Labels \cite{HollandEtAl2018}, Data Statements \cite{statements}, or Factsheets \cite{factsheets}, describing datasets that the model was trained and evaluated on. Model cards provide a way to inform users about what machine learning systems can and cannot do, the types of errors they make, and additional steps that could create more fair and inclusive outcomes with the technology.

\section{Background}

Many mature industries have developed standardized methods of benchmarking various systems under different conditions. 
For example, as noted in~\cite{GebruEtAl2018}, the electronic hardware industry provides datasheets with detailed characterizations of components' performances under different test conditions. By contrast, despite the broad reach and impact of machine learning models, there are no standard stress tests that are performed on machine learning based systems, nor standardized formats to report the results of these tests. Recently, researchers have proposed standardized forms of communicating characteristics of datasets used in machine learning~\cite{GebruEtAl2018,HollandEtAl2018, statements} to help users understand the context in which the datasets should be used. 
We focus on the complementary task for machine learning models, proposing a standardized method to evaluate the performance of human-centric models:  Disaggregated by unitary and intersectional groups such as cultural, demographic, or phenotypic population groups.  A framework that we refer to as ``Model Cards'' can present such evaluation supplemented with additional considerations such as intended use.

Outside of machine learning, the need for population-based reporting of outcomes as suggested here has become increasingly evident. For example, in vehicular crash tests, dummies with prototypical female characteristics were only introduced after researchers discovered that women were more likely than men to suffer serious head injuries in real-world side impacts \cite{IIHS2003}. 
Similarly, drugs developed based on results of clinical trials with exclusively male participants have led to overdosing in women~ \cite{GuardianFDA,FDA2013}. In 1998, the U.S. Food and Drug Administration mandated that clinical trial results be disaggregated by groups such as age, race and gender~\cite{fda1998}.  

While population-based analyses of errors and successes can be provided for unitary groups such as ``men'', ``women'', and ``non-binary'' gender groups, they should also be provided intersectionally, looking at two or more characteristics such as gender and age simultaneously. Intersectional analyses are linked to intersectionality theory, which describes how discrete experiences associated with characteristics like race or gender in isolation do not accurately reflect their interaction~\cite{intersection}. Kimberl\'e Crenshaw, who pioneered intersectional research in critical race theory, discusses the story of Emma DeGraffenreid, who was part of a failed lawsuit against General Motors in 1976, claiming that the company's hiring practices discriminated against Black women. In their court opinion, the judges noted that since General Motors hired many women for secretarial positions, and many Black people for factory roles, they could not have discriminated against Black women. However, what the courts failed to see was that only White women were hired into secretarial positions and only Black men were hired into factory roles. Thus, Black women like Emma DeGraffenreid had no chance of being employed at General Motors. This example highlights the importance of intersectional analyses: empirical analyses that emphasize the interaction between various demographic categories including race, gender, and age.

Before further discussing the details of the model card, it is important to note that at least two of the three characteristics discussed so far, race and gender, are socially sensitive.  Although analyzing models by race and gender may follow from intersectionality theory, how ``ground truth'' race or gender categories should be labeled in a dataset, and whether or not datasets should be labeled with these categories at all, is not always clear. This issue is further confounded by the complex relationship between gender and sex.  When using cultural identity categories such as race and gender to subdivide analyses, and depending on the context, we recommend either using datasets with self-identified labels or with labels clearly designated as {\it perceived} (rather than self-identified).  When this is not possible, datasets of public figures with known public identity labels may be useful.  Further research is necessary to expand how groups may be defined, for example, by automatically discovering groups with similarities in the evaluation datasets.

\section{Motivation}\label{sec:motivation}

As the use of machine learning technology has rapidly increased, so too have reports of errors and failures. 
Despite the potentially serious repercussions of these errors, those looking to use trained machine learning models in a particular context have no way of understanding the systematic impacts of these models before deploying them.

The proposal of ``Model Cards'' specifically aims to standardize ethical practice and reporting - allowing stakeholders to compare candidate models for deployment across not only traditional evaluation metrics but also along the axes of ethical, inclusive, and fair considerations. This goes further than current solutions to aid stakeholders in different contexts.  For example, to aid policy makers and regulators on questions to ask of a model, and known benchmarks around the suitability of a model in a given setting. 

Model reporting will hold different meaning to those involved in different aspects of model development, deployment, and use.  Below, we outline a few use cases for different stakeholders:
\begin{itemize}
 \item {\bf ML and AI practitioners} can better understand how well the model might work for the intended use cases and track its performance over time.
\item {\bf Model developers} can compare the model's results to other models in the same space, and make decisions about training their own system.
\item {\bf Software developers} working on products that use the model's predictions can inform their design and implementation decisions.
\item {\bf Policymakers} can understand how a machine learning system may fail or succeed in ways that impact people.
\item {\bf Organizations} can inform decisions about adopting technology that incorporates machine learning.
\item {\bf ML-knowledgeable individuals} can be informed on different options for fine-tuning, model combination, or additional rules and constraints to help curate models for intended use cases without requiring technical expertise. 
\item {\bf Impacted individuals} who may experience effects from a model can better understand how it works or use information in the card to pursue remedies.
\end{itemize}

Not only does this practice improve model understanding and help to standardize decision making processes for invested stakeholders, but it also encourages forward-looking model analysis techniques. For example, slicing the evaluation across groups functions to highlight errors that may fall disproportionately on some groups of people, and accords with many recent notions of mathematical fairness (discussed further in the example model card in Figure \ref{fig:smiling_card}).  Including group analysis as part of the reporting procedure prepares stakeholders to begin to gauge the fairness and inclusion of future outcomes of the machine learning system. Thus, in addition to supporting decision-making processes for determining the suitability of a given machine learning model in a particular context,  model reporting is an approach for responsible transparent and accountable practices in machine learning.

People and organizations releasing models may be additionally incentivized to provide model card details because it helps potential users of the models to be better informed on which models are best for their specific purposes.  If model card reporting becomes standard, potential users can compare and contrast different models in a well-informed way.  Results on several different evaluation datasets will additionally aid potential users, although evaluation datasets suitable for disaggregated evaluation are not yet common.  Future research could include creating robust evaluation datasets and protocols for the types of disaggregated evaluation we advocate for in this work, for example, by including differential privacy mechanisms \cite{DworkDifferential} so that individuals in the testing set cannot be uniquely identified by their characteristics.

\section{Model Card Sections}

Model cards serve to disclose information about a trained machine learning model. This includes how it was built, what assumptions were made during its development, what type of model behavior different cultural, demographic, or phenotypic population groups may experience, and an evaluation of how well the model performs with respect to those groups. Here, we propose a set of sections that a model card should have, and details that can inform the stakeholders discussed in Section \ref{sec:motivation}.  A summary of all suggested sections is provided in Figure \ref{fig:model_sections}.

The proposed set of sections below are intended to provide relevant details to consider, but are not intended to be complete or exhaustive, and may be tailored depending on the model, context, and stakeholders.  Additional details may include, for example, interpretability approaches, such as saliency maps, TCAV \cite{KimEtAl2018}, and Path-Integrated Gradients \cite{PathIntegratedGradients2017,PathIntegratedGradients2018}); stakeholder-relevant explanations (e.g., informed by a careful consideration of philosophical, psychological, and other factors concerning what is as a good explanation in different contexts \cite{GoogleResponsiblePractices}); and privacy approaches used in model training and serving.

\begin{figure}
\begin{framed}
{\Large {\bf Model Card}}
\begin{itemize}[leftmargin=*]
\item {\bf Model Details}.  Basic information about the model.
\begin{itemize}
\item Person or organization developing model
\item Model date
\item Model version
\item Model type
\item Information about training algorithms, parameters, fairness constraints or other applied approaches, and features
\item Paper or other resource for more information
\item Citation details
\item License
\item Where to send questions or comments about the model \end{itemize}
\item {\bf Intended Use}.  Use cases that were envisioned during development. 
\begin{itemize}
\item Primary intended uses
\item Primary intended users
\item Out-of-scope use cases
\end{itemize}
\item {\bf Factors}.  Factors could include demographic or phenotypic groups, environmental conditions, technical attributes, or others listed in Section \ref{sec:factors}.  
\begin{itemize}
\item Relevant factors
\item Evaluation factors
\end{itemize}
\item {\bf Metrics}.  Metrics should be chosen to reflect potential real-world impacts of the model.
\begin{itemize}
\item Model performance measures
\item Decision thresholds
\item Variation approaches
\end{itemize}
\item {\bf Evaluation Data}.  Details on the dataset(s) used for the quantitative analyses in the card.
\begin{itemize}
\item Datasets
\item Motivation
\item Preprocessing
\end{itemize}
\item {\bf Training Data}.  May not be possible to provide in practice.  When possible, this section should mirror Evaluation Data.  If such detail is not possible, minimal allowable information should be provided here, such as details of the distribution over various factors in the training datasets.
\item {\bf Quantitative Analyses}
\begin{itemize}
\item Unitary results
\item Intersectional results
\end{itemize}
\item {\bf Ethical Considerations}
\item {\bf Caveats and Recommendations}
\vspace{-.25em}
\end{itemize}
\end{framed}
\vspace{-1em}
\caption{Summary of model card sections and suggested prompts for each.}\label{fig:model_sections}
\vspace{-1em}
\end{figure}
\subsection{Model Details}
This section of the model card should serve to answer basic questions regarding the model version, type and other details.  \\
{\bf Person or organization developing model}:  What person or organization developed the model? This can be used by all stakeholders to infer details pertaining to model development and potential conflicts of interest. \\
{\bf Model date}:  When was the model developed?  This is useful for all stakeholders to become further informed on what techniques and data sources were likely to be available during model development. \\
{\bf Model version}:  Which version of the model is it, and how does it differ from previous versions?  This is useful for all stakeholders to track whether the model is the latest version, associate known bugs to the correct model versions, and aid in model comparisons. \\
{\bf Model type}:   What type of model is it?  This includes basic model architecture details, such as whether it is a  Naive Bayes classifier, a Convolutional Neural Network, etc.  This is likely to be particularly relevant for software and model developers, as well as individuals knowledgeable about machine learning, to highlight what kinds of assumptions are encoded in the system.\\
{\bf Paper or other resource for more information}: Where can resources for more information be found?  \\
{\bf Citation details}: How should the model be cited? \\
{\bf License}: License information can be provided.  \\
{\bf Feedback on the model}:  E.g., what is an email address that people may write to for further information? \\

There are cases where some of this information may be sensitive. For example, the amount of detail corporations choose to disclose might be different from academic research groups. This section should not be seen as a requirement to compromise private information or reveal proprietary training techniques; rather, a place to disclose basic decisions and facts about the model that the organization can share with the broader community in order to better inform on what the model represents.

\subsection{Intended Use}
This section should allow readers to quickly grasp what the model should and should not be used for, and why it was created. 
It can also help frame the statistical analysis presented in the rest of the card, including a short description of the user(s), use-case(s), and context(s) for which the model was originally developed. Possible information includes: \\
{\bf Primary intended uses}: This section details whether the model was developed with general or specific tasks in mind (e.g., plant recognition worldwide or in the Pacific Northwest). The use cases may be as broadly or narrowly defined as the developers intend. For example, if the model was built simply to label images, then this task should be indicated as the primary intended use case. \\
{\bf Primary intended users}:  For example, was the model developed for entertainment purposes, for hobbyists, or enterprise solutions?  This helps users gain insight into how robust the model may be to different kinds of inputs. \\
{\bf Out-of-scope uses}: 
Here, the model card should highlight technology that the model might easily be confused with, or related contexts that users could try to apply the model to. This section may provide an opportunity to recommend a related or similar model that was designed to better meet that particular need, where possible. This section is inspired by warning labels on food and toys, and similar disclaimers presented in electronic datasheets. Examples include ``not for use on text examples shorter than 100 tokens'' or ``for use on black-and-white images only; please consider our research group's full-color-image classifier for color images.''

\subsection{Factors}\label{sec:factors}

Model cards ideally provide a summary of model performance across a variety of relevant factors including {\it groups}, {\it instrumentation}, and {\it environments}. We briefly describe each of these factors and their relevance followed by the corresponding prompts in the model card.

\subsubsection{Groups} ``Groups'' refers to distinct categories with similar characteristics that are present in the evaluation data instances. For human-centric machine learning models, ``groups'' are people who share one or multiple characteristics.  
Intersectional model analysis for human-centric models is inspired by the sociological concept of intersectionality, which explores how an individual's identity and experiences are shaped not just by unitary personal characteristics -- such as race, gender, sexual orientation or health -- but instead by a complex combination of many factors. These characteristics, which include but are not limited to cultural, demographic and phenotypic categories, are important to consider when evaluating machine learning models. Determining which groups to include in an intersectional analysis requires examining the intended use of the model and the context under which it may be deployed. Depending on the situation, certain groups may be more vulnerable than others to unjust or prejudicial treatment.

For human-centric computer vision models, the visual presentation of age, gender, and Fitzpatrick skin type~\cite{fitzpatrick} may be relevant. However, this must be balanced with the goal of preserving the privacy of individuals. As such, collaboration with policy, privacy, and legal experts is necessary in order to ascertain which groups may be responsibly inferred, and how that information should be stored and accessed (for example, using differential privacy \cite{DworkDifferential}).

Details pertaining to groups, including who annotated the training and evaluation datasets, instructions and compensation given to annotators, and inter-annotator agreement, should be provided as part of the data documentation made available with the dataset.  See \cite{GebruEtAl2018,HollandEtAl2018,statements} for more details.

\subsubsection{Instrumentation}
In addition to groups, the performance of a model can vary depending on what instruments were used to capture the input to the model. For example, a face detection model may perform differently depending on the camera's hardware and software, including lens, image stabilization, high dynamic range techniques, and background blurring for portrait mode. Performance may also vary across real or simulated traditional camera settings such as aperture, shutter speed and ISO. Similarly, video and audio input will be dependent on the choice of recording instruments and their settings. 
\subsubsection{Environment}
A further factor affecting model performance is the environment in which it is deployed.  For example, face detection systems are often less accurate under low lighting conditions or when the air is humid \cite{light}. Specifications across different lighting and moisture conditions would help users understand the impacts of these environmental factors on model performance.

\subsubsection{Card Prompts}
We propose that the Factors section of model cards expands on two prompts: \\
\noindent{\bf Relevant factors}:  What are foreseeable salient factors for which model performance may vary, and how were these determined? \\
{\bf Evaluation factors}:  Which factors are being reported, and why were these chosen?  If the relevant factors and evaluation factors are different, why? For example, while Fitzpatrick skin type is a relevant factor for face detection, an evaluation dataset annotated by skin type might not be available until reporting model performance across groups becomes standard practice.

\subsection{Metrics}\label{sec:metrics}
The appropriate metrics to feature in a model card depend on the type of model that is being tested. For example, classification systems in which the primary output is a class label differ significantly from systems whose primary output is a score.
In all cases, the reported metrics should be determined based on the model's structure and intended use. Details for this section include: \\
{\bf Model performance measures}:  What measures of model performance are being reported, and why were they selected over other measures of model performance?   \\
{\bf Decision thresholds}:  If decision thresholds are used, what are they, and why were those decision thresholds chosen?  When the model card is presented in a digital format, a threshold slider should ideally be available to view performance parameters across various decision thresholds.\\
{\bf Approaches to uncertainty and variability}:  How are the measurements and estimations of these metrics calculated? For example, this may include standard deviation, variance, confidence intervals, or KL divergence. Details of how these values are approximated should also be included (e.g., average of 5 runs, 10-fold cross-validation).
\subsubsection{Classification systems}
For classification systems, the error types that can be derived from a confusion matrix are {\it false positive rate}, {\it false negative rate}, {\it false discovery rate}, and {\it false omission rate}. We note that the relative importance of each of these metrics is system, product and context dependent. 

For example, in a surveillance scenario, surveillors may value a low false negative rate (or the rate at which the surveillance system fails to detect a person or an object when it should have). On the other hand, those being surveilled may value a low false positive rate (or the rate at which the surveillance system detects a person or an object when it should not have). 
We recommend listing all values and providing context about which were prioritized during development and why.

Equality between some of the different confusion matrix metrics is equivalent to some definitions of fairness. For example, equal false negative rates across groups is equivalent to fulfilling Equality of Opportunity, and equal false negative and false positive rates across groups is equivalent to fulfilling Equality of Odds \cite{HardtEtAl2016}.

\subsubsection{Score-based analyses}
For score-based systems such as pricing models and risk assessment algorithms, describing differences in the distribution of measured metrics across groups may be helpful. For example, reporting measures of central tendency such as the mode, median and mean, as well as measures of dispersion or variation such as the range, quartiles, absolute deviation, variance and standard deviation could facilitate the statistical commentary necessary to make more informed decisions about model development. A model card could even extend beyond these summary statistics to reveal other measures of differences between distributions such as cross entropy, perplexity, KL divergence and pinned area under the curve (pinned AUC)~\cite{DixonEtAl2018}.

There are a number of applications that do not appear to be score-based at first glance, but can be considered as such for the purposes of intersectional analysis. For instance, a model card for a translation system could compare BLEU scores~\cite{bleu} across demographic groups, and a model card for a speech recognition system could compare word-error rates. Although the primary outputs of these systems are not scores, looking at the score differences between populations may yield meaningful insights since comparing raw inputs quickly grows too complex. 

\subsubsection{Confidence}
Performance metrics that are disaggregated by various combinations of instrumentation, environments and groups makes it especially important to understand the confidence intervals for the reported metrics. Confidence intervals for metrics derived from confusion matrices can be calculated by treating the matrices as probabilistic models of system performance \cite{goutte2005probabilistic}.

\subsection{Evaluation Data}
All referenced datasets would ideally point to any set of documents that provide visibility into the source and composition of the dataset. Evaluation datasets should include datasets that are publicly available for third-party use. These could be existing datasets or new ones provided alongside the model card analyses to enable further benchmarking.  Potential details include: \\
{\bf Datasets}:  What datasets were used to evaluate the model?   \\
{\bf Motivation}:  Why were these datasets chosen?\\
{\bf Preprocessing}:  How was the data preprocessed for evaluation (e.g., tokenization of sentences, cropping of images, any filtering such as dropping images without faces)?  \\

To ensure that model cards are statistically accurate and verifiable, the evaluation datasets should not only be representative of the model's typical use cases but also anticipated test scenarios and challenging cases.  For instance, if a model is intended for use in a workplace that is phenotypically and demographically homogeneous, and trained on a dataset that is representative of the expected use case, it may be valuable to evaluate that model on two evaluation sets: one that matches the workplace's population, and another set that contains individuals that might be more challenging for the model (such as children, the elderly, and people from outside the typical workplace population). This methodology can highlight pathological issues that may not be evident in more routine testing. 

It is often difficult to find datasets that represent populations outside of the initial domain used in training. In some of these situations, synthetically generated datasets 
may provide representation for use cases that would otherwise go unevaluated~\cite{synthetic}. Section \ref{toxicity} provides an example of including synthetic data in the model evaluation dataset.

\subsection{Training Data}
Ideally, the model card would contain as much information about the training data as the evaluation data. However, there might be cases where it is not feasible to provide this level of detailed information about the training data. For example, the data may be proprietary, or require a non-disclosure agreement. In these cases, we advocate for basic details about the distributions over groups in the data, as well as any other details that could inform stakeholders on the kinds of biases the model may have encoded.
\subsection{Quantitative Analyses}
Quantitative analyses should be {\it disaggregated}, that is, broken down by the chosen factors.  Quantitative analyses should provide the results of evaluating the model according to the chosen metrics, providing confidence interval values when possible.  Parity on the different metrics across disaggregated population subgroups corresponds to how {\it fairness} is often defined \cite{verma2018fairness,MitchellEtAl2018}.  Quantitative analyses should demonstrate the metric variation (e.g., with error bars), as discussed in Section \ref{sec:metrics} and visualized in Figure \ref{fig:smiling_card}.  

\noindent The disaggregated evaluation includes: \\
{\bf Unitary results}:  How did the model perform with respect to each factor? \\
{\bf Intersectional results}:  How did the model perform with respect to the intersection of evaluated factors?

\subsection{Ethical Considerations}

This section is intended to demonstrate the ethical considerations that went into model development, surfacing ethical challenges and solutions to stakeholders.  Ethical analysis does not always lead to precise solutions, but the process of ethical contemplation is worthwhile to inform on responsible practices and next steps in future work.

While there are many frameworks for ethical decision-making in technology that can be adapted here \cite{markkula,deon,ethicalos}, the following are specific questions you may want to explore in this section: \\
{\bf Data}: Does the model use any sensitive data (e.g., protected classes)? \\
{\bf Human life}: Is the model intended to inform decisions about matters central to human life or flourishing -- e.g., health or safety?  Or could it be used in such a way? \\
{\bf Mitigations}:  What risk mitigation strategies were used during model development? \\
{\bf Risks and harms}: What risks may be present in model usage? Try to identify the potential recipients, likelihood, and magnitude of harms. If these cannot be determined, note that they were considered but remain unknown. \\
{\bf Use cases}: Are there any known model use cases that are especially fraught? This may connect directly to the intended use section of the model card.

If possible, this section should also include any additional ethical considerations that went into model development, for example, review by an external board, or testing with a specific community.

\subsection{Caveats and Recommendations}
This section should list additional concerns that were not covered in the previous sections. For example, did the results suggest any further testing?  Were there any relevant groups that were not represented in the evaluation dataset?  Are there additional recommendations for model use?  What are the ideal characteristics of an evaluation dataset for this model?

\section{Examples}\label{sec:examples}
We present worked examples of model cards for two models: an image-based classification system and a text-based scoring system. 

\subsection{Smiling Classifier}
\begin{figure*}
\raggedright
\begin{framed}
\begin{center}{\LARGE {\bf Model Card - Smiling Detection in Images}}\end{center} 
\vspace{1em}
    \begin{minipage}{.615\textwidth}
{\bf Model Details}
\begin{itemize}[leftmargin=*]
\item Developed by researchers at Google and the University of Toronto, 2018, v1.
\item Convolutional Neural Net.
\item Pretrained for face recognition then fine-tuned with cross-entropy loss for binary smiling classification.
\end{itemize}

\vspace{.25em}
{\bf Intended Use}
\begin{itemize}[leftmargin=*]
\item Intended to be used for fun applications, such as creating cartoon smiles on real images; augmentative applications, such as providing details for people who are blind; or assisting applications such as automatically finding smiling photos.
\item Particularly intended for younger audiences. 
\item Not suitable for emotion detection or determining affect; smiles were annotated based on physical appearance, and not underlying emotions.
\end{itemize}

\vspace{.25em}
{\bf Factors}
\begin{itemize}[leftmargin=*]
\item Based on known problems with computer vision face technology, potential relevant factors include groups for gender, age, race, and Fitzpatrick skin type; hardware factors of camera type and lens type; and environmental factors of lighting and humidity.
\item Evaluation factors are gender and age group, as annotated in the publicly available dataset CelebA \cite{LiuEtAl2015}.  Further possible factors not currently available in a public smiling dataset.  Gender and age determined by third-party annotators based on visual presentation, following a set of examples of male/female gender and young/old age.  Further details available in \cite{LiuEtAl2015}.
\end{itemize}

\vspace{.25em}
{\bf Metrics}
\begin{itemize}[leftmargin=*]
\item Evaluation metrics include {\bf False Positive Rate} and {\bf False Negative Rate} to measure disproportionate model performance errors across subgroups. {\bf False Discovery Rate} and {\bf False Omission Rate}, which measure the fraction of negative (not smiling) and positive (smiling) predictions that are incorrectly predicted to be positive and negative, respectively, are also reported. \cite{verma2018fairness}  
\item Together, these four metrics provide values for different errors that can be calculated from the confusion matrix for binary classification systems.
\item These also correspond to metrics in recent definitions of ``fairness'' in machine learning (cf. \cite{HardtEtAl2016,chouldechova2017fair}), where parity across subgroups for different metrics correspond to different fairness criteria.
\item 95\% confidence intervals calculated with bootstrap resampling.  
\item All metrics reported at the .5 decision threshold, where all error types (FPR, FNR, FDR, FOR) are within the same range (0.04 - 0.14). 
\end{itemize}
\vspace{.3em}
\begin{minipage}{0.4\textwidth}
{\bf Training Data}
\begin{itemize}[leftmargin=*]
\item CelebA \cite{LiuEtAl2015}, training data split. 
\end{itemize}
\vspace{1em}
\end{minipage}
\hspace{2em}
\begin{minipage}{0.5\textwidth}
    {\bf Evaluation Data}
\begin{itemize}[leftmargin=*]
\item CelebA \cite{LiuEtAl2015}, test data split.
\item Chosen as a basic proof-of-concept.
\end{itemize}
\end{minipage}

\vspace{-.25em}
{\bf Ethical Considerations}
\begin{itemize}[leftmargin=*]
\item Faces and annotations based on public figures (celebrities). No new information is inferred or annotated.
\end{itemize}
\end{minipage}
\begin{minipage}{0.31\textwidth}
\vspace{.5em}
{\bf Quantitative Analyses} \\
\begin{subfigure}{\textwidth}
\vspace{.25em}
\centering
\hbox{\includegraphics[scale=0.6,trim={5.1cm 0 0 0},clip]
{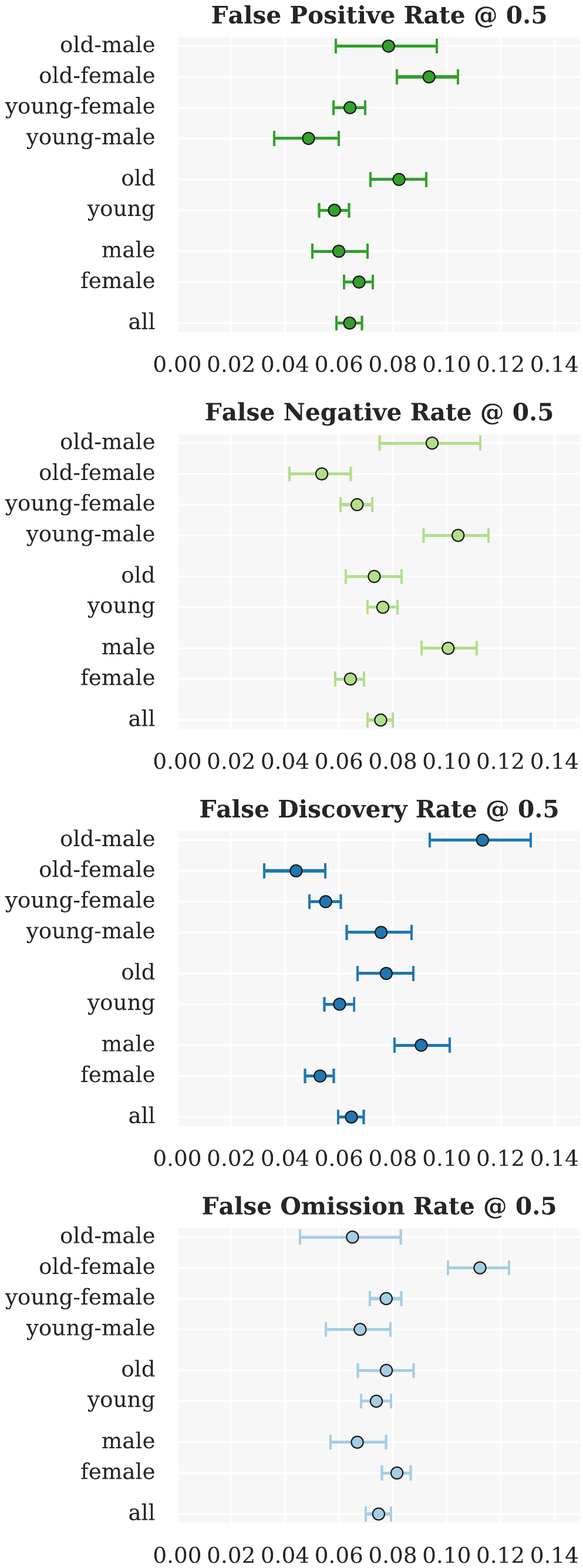}}
\label{fig:celeba_error_rates_050}
\end{subfigure}
    \end{minipage}
\vspace{-.15em}    

{\bf Caveats and Recommendations}
\begin{itemize}[leftmargin=*]
\item Does not capture race or skin type, which has been reported as a source of disproportionate errors \cite{BuolamwiniGebru2018}.
\item Given gender classes are binary (male/not male), which we include as male/female. Further work needed to evaluate across a spectrum of genders.
\item An ideal evaluation dataset would additionally include annotations for Fitzpatrick skin type, camera details, and environment (lighting/humidity) details.
\end{itemize}    
\end{framed}
\vspace{-1em}
\caption{Example Model Card for a smile detector trained and evaluated on the CelebA dataset.}\label{fig:smiling_card}
\end{figure*}

To show an example of a model card for an image classification problem, we use the public CelebA dataset~\cite{LiuEtAl2015} to examine the performance of a trained ``smiling'' classifier across both age and gender categories. Figure \ref{fig:smiling_card} shows our prototype.

These results demonstrate a few potential issues. 
For example, the false discovery rate on older men is much higher than that for other groups. This means that many predictions incorrectly classify older men as smiling when they are not. On the other hand, men (in aggregate) have a higher false negative rate, meaning that many of the men that are in fact smiling in the photos are incorrectly classified as not smiling. 

The results of these analyses give insight into contexts the model might not be best suited for. For example, it may not be advisable to apply the model on a diverse group of audiences, and it may be the most useful when detecting the presence of a smile is more important than detecting its absence (for example, in an application that automatically finds `fun moments' in images).  Additional fine-tuning, for example, with images of older men, may help create a more balanced performance across groups.

\subsection{Toxicity Scoring} \label{toxicity}

\begin{figure*}
\raggedright
\begin{framed}
\begin{center}{\LARGE {\bf Model Card - Toxicity in Text}}\end{center} 
    \begin{minipage}{.45\textwidth}

{\bf Model Details} 
\begin{itemize}[leftmargin=*]
\item The TOXICITY classifier provided by Perspective API \cite{perspective_api}, trained to predict the likelihood that a comment will be perceived as toxic. 
\item Convolutional Neural Network.
\item Developed by Jigsaw in 2017.
\end{itemize}

{\bf Intended Use}
\begin{itemize}[leftmargin=*]
\item Intended to be used for a wide range of use cases such as supporting human moderation and providing feedback to comment authors.
\item Not intended for fully automated moderation.
\item Not intended to make judgments about specific individuals.
\end{itemize}

{\bf Factors}
\begin{itemize}[leftmargin=*]
\item Identity terms referencing frequently attacked groups, focusing on sexual orientation, gender identity, and race.
\end{itemize}

{\bf Metrics}
\begin{itemize}[leftmargin=*]
\item Pinned AUC, as presented in \cite{DixonEtAl2018}, which measures threshold-agnostic separability of toxic and non-toxic comments for each group, within the context of a background distribution of other groups.
\end{itemize}

{\bf Ethical Considerations}
\begin{itemize}[leftmargin=*]
\item Following \cite{conversation_ai}, the Perspective API uses a set of values to guide their work. These values are Community, Transparency, Inclusivity, Privacy, and Topic-neutrality. Because of privacy considerations, the model does not take into account user history when making judgments about toxicity.
\end{itemize}
\vspace{1em}
 \end{minipage}
 \hspace{2em}
     \begin{minipage}{.45\textwidth}
\vspace{2em}
{\bf Training Data} 
\begin{itemize}[leftmargin=*]
\item Proprietary from Perspective API.  Following details in \cite{DixonEtAl2018} and \cite{perspective_api}, this includes comments from a online forums such as Wikipedia and New York Times, with crowdsourced labels of whether the comment is ``toxic''.
\item ``Toxic'' is defined as ``a rude, disrespectful, or unreasonable comment that is likely to make you leave a discussion.''
\end{itemize}

{\bf Evaluation Data} 
\begin{itemize}[leftmargin=*]
\item A synthetic test set generated using a template-based approach, as suggested in \cite{DixonEtAl2018}, where identity terms are swapped into a variety of template sentences.
\item Synthetic data is valuable here because \cite{DixonEtAl2018} shows that real data often has disproportionate amounts of toxicity directed at specific groups. Synthetic data ensures that we evaluate on data that represents both toxic and non-toxic statements referencing a variety of groups. 
\end{itemize}

{\bf Caveats and Recommendations}
\begin{itemize}[leftmargin=*]
\item Synthetic test data covers only a small set of very specific comments. While these are designed to be representative of common use cases and concerns, it is not comprehensive.
\end{itemize}
\vspace{11em}
\end{minipage}
{\bf Quantitative Analyses} \\
\hspace{-.65em}\begin{subfigure}{\textwidth}
\centering
\begin{tabular}{c@{}c}
\includegraphics[scale=0.26,trim={0 20.1cm 0 0},clip]{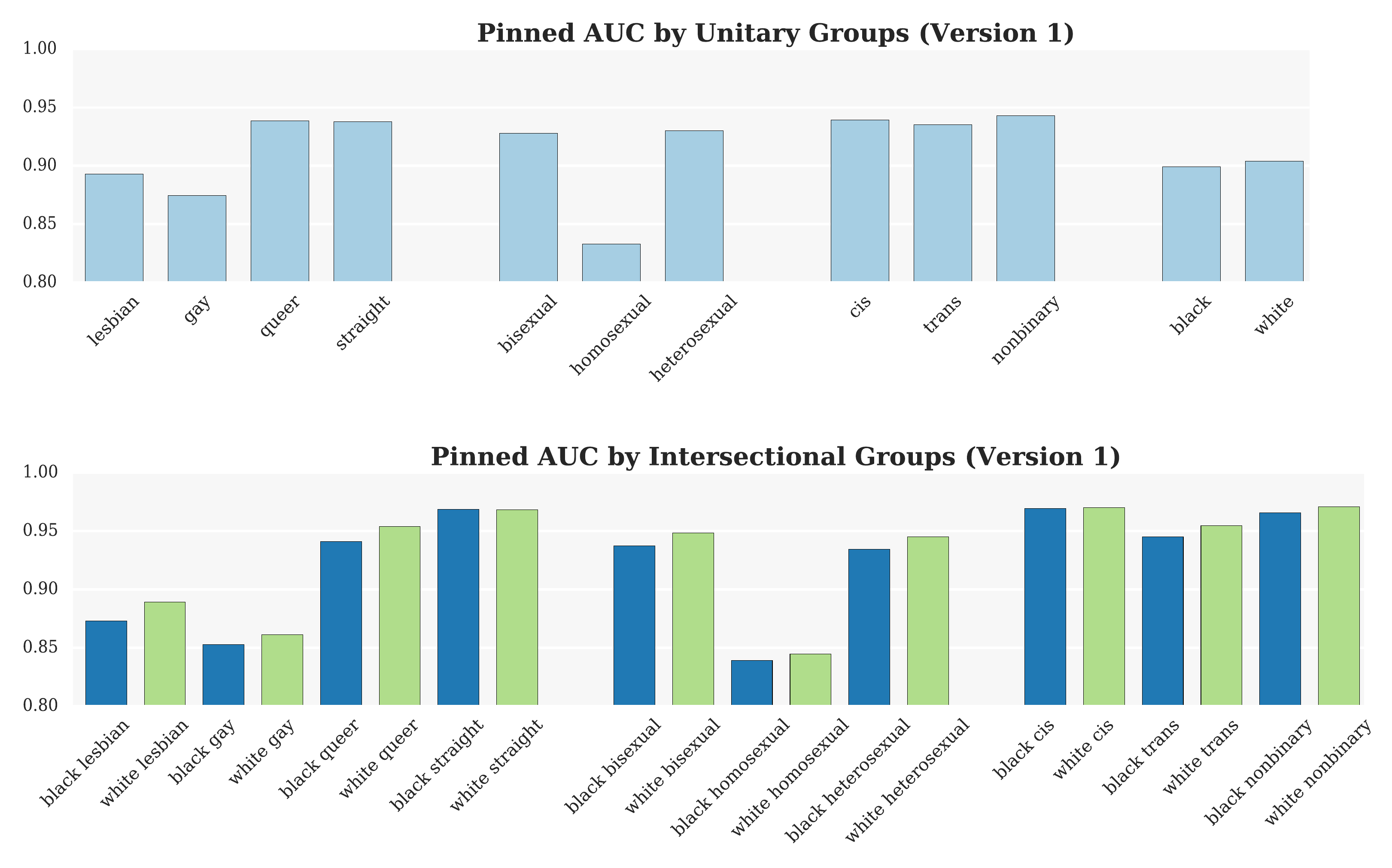} & \vspace{.1em}\hspace{-.2em}\includegraphics[scale=0.26,trim={0 0 0  20cm},clip]{pinned_auc_toxicity.pdf}
\end{tabular}

\end{subfigure}


\end{framed}
\caption{Example Model Card for two versions of Perspective API's toxicity detector.}\label{fig:toxicity_card}\label{fig:toxicity_analysis}
\end{figure*}

Our second example provides a model card for Perspective API's TOXICITY classifier built to detect `toxicity' in text \cite{perspective_api}, and is presented in Figure \ref{fig:toxicity_card}. To evaluate the model, we use an intersectional version of the open source, synthetically created Identity Phrase Templates test set published in \cite{DixonEtAl2018}. We show two versions of the quantitative analysis: one for TOXICITY v.~1, the initial version of the this model, and one for TOXICITY v.~5, the latest version.

This model card highlights the drastic ways that models can change over time, and the importance of having a model card that is updated with each new model release. TOXICITY v.~1 has low performance for several terms, especially ``lesbian'', ``gay'', and ``homosexual''. This is consistent with what some users of the initial TOXICITY model found, as reported by the team behind Perspective API in \cite{false_positive_blog}. Also in \cite{false_positive_blog}, the Perspective API team shares the bias mitigation techniques they applied to the TOXICITY v.~1 model, in order to create the more equitable performance in TOXICITY v.~5. By making model cards a standard part of API launches, teams like the Perspective API team may be able to find and mitigate some of these biases earlier.


\section{Discussion \& Future Work}

We have proposed frameworks called model cards for reporting information about what a trained machine learning model is and how well it works.  Model cards include information about the context of the model, as well as model performance results disaggregated by different unitary and intersectional population groups.  Model cards are intended to accompany a model after careful review has determined that the foreseeable benefits outweigh the foreseeable risks in the model's use or release.  

To demonstrate the use of model cards in practice, we have provided two examples: A model card for a smiling classifier tested on the CelebA dataset, and a model card for a public toxicity detector tested on the Identity Phrase Templates dataset. We report confusion matrix metrics for the smile classifier and Pinned AUC for the toxicity detector, along with model details, intended use, pointers to information about training and evaluation data, ethical considerations, and further caveats and recommendations. 

The framework presented here is intended to be general enough to be applicable across different institutions, contexts, and stakeholders.  It also is suitable for recently proposed requirements for analysis of algorithmic decision systems in critical social institutions, for example, for models used in determining government benefits, employment evaluations, criminal risk assessment, and criminal DNA analysis \cite{LitigatingAlgorithms2018}.

Model cards are just one approach to increasing transparency between developers, users, and stakeholders of machine learning models and systems. They are designed to be flexible in both scope and specificity in order to accommodate the wide variety of machine learning model types and potential use cases. Therefore the usefulness and accuracy of a model card relies on the integrity of the creator(s) of the card itself. It seems unlikely, at least in the near term, that model cards could be standardized or formalized to a degree needed to prevent misleading representations of model results (whether intended or unintended). It is therefore important to consider model cards as one transparency tool among many, which could include, for example, algorithmic auditing by third-parties (both quantitative and qualitative), ``adversarial testing'' by technical and non-technical analysts, and more inclusive user feedback mechanisms. Future work will aim to refine the methodology of creating model cards by studying how model information is interpreted and used by different stakeholders. Researchers should also explore how model cards can strengthen and complement other transparency methods
\section{Acknowledgements}
Thank you to Joy Buolamwini, Shalini Ananda and Shira Mitchell for invaluable conversations and insight.

\bibliographystyle{ACM-Reference-Format}
\bibliography{references}

\appendix

\end{document}